\documentclass[pdflatex,sn-mathphys-ay,iicol]{sn-jnl}

\usepackage{graphicx}%
\usepackage{multirow}%
\usepackage{amsmath,amssymb,amsfonts}%
\usepackage{amsthm}%
\usepackage{mathrsfs}%
\usepackage[title]{appendix}%
\usepackage{xcolor}%
\usepackage{textcomp}%
\usepackage{manyfoot}%
\usepackage{booktabs}%
\usepackage{algorithm}%
\usepackage{algorithmicx}%
\usepackage{algpseudocode}%
\usepackage{listings}%
\usepackage{makecell}
\usepackage{float}      
\usepackage{graphicx}  
\usepackage{hyperref}
\usepackage{tabularx}
\usepackage{natbib}

\theoremstyle{thmstyleone}%
\theoremstyle{thmstyletwo}%

\theoremstyle{thmstylethree}%

\begin{document}

\title[ST-GDance++: A Scalable Spatial–Temporal Diffusion for Long-Duration Group Choreography]{ST-GDance++: A Scalable Spatial–Temporal Diffusion for Long-Duration Group Choreography}


\author[1]{\fnm{Jing} \sur{Xu}}\email{jing.xu1@monash.edu}
\author*[1]{\fnm{Weiqiang} \sur{Wang}}\email{weiqiang.wang@monash.edu}
\author*[1]{\fnm{Cunjian} \sur{Chen}}\email{cunjian.chen@monash.edu}
\author[2]{\fnm{Jun} \sur{Liu}}\email{j.liu81@lancaster.ac.uk}
\author[1]{\fnm{Qiuhong} \sur{Ke}}\email{qiuhong.ke@monash.edu}


\affil[1]{%
\orgdiv{Faculty of Information Technology},
\orgname{Monash University},
\orgaddress{
\city{Melbourne},
\state{VIC},
\postcode{3800},
\country{Australia}
}}

\affil[2]{%
\orgdiv{School of Computing and Communications},
\orgname{Lancaster University},
\orgaddress{
\city{Lancaster},
\postcode{LA1 4YW},
\country{UK}
}}



\abstract{Group dance generation from music requires synchronizing multiple dancers while maintaining spatial coordination, making it highly relevant to applications such as film production, gaming, and animation. Recent group dance generation models have achieved promising generation quality, but they remain difficult to deploy in interactive scenarios due to bidirectional attention dependencies. As the number of dancers and the sequence length increase, the attention computation required for aligning music conditions with motion sequences grows quadratically, leading to reduced efficiency and increased risk of motion collisions. Effectively modeling dense spatial–temporal interactions is therefore essential, yet existing methods often struggle to capture such complexity, resulting in limited scalability and unstable multi-dancer coordination. To address these challenges, we propose ST-GDance++, a scalable framework that decouples spatial and temporal dependencies to enable efficient and collision-aware group choreography generation. For spatial modeling, we introduce lightweight distance-aware graph convolutions to capture inter-dancer relationships while reducing computational overhead. For temporal modeling, we design a diffusion noise scheduling strategy together with an efficient temporal-aligned attention mask, enabling stream-based generation for long motion sequences and improving scalability in long-duration scenarios. Experiments on the AIOZ-GDance dataset show that ST-GDance++ achieves competitive generation quality with significantly reduced latency compared to existing methods. \href{https://yilliajing.github.io/ST-GDancepp-Website/}{Project page}.}

\keywords{Music-driven Dance Generation, Group Choreography, Spatial–Temporal Modeling, Diffusion Model}



\maketitle

\section{Introduction}\label{intro}

Dance is a highly expressive art form that uses body movements to convey emotion. With the rise of generative AI, music-to-dance generation has found increasing applications in film, cultural productions, gaming, and beyond~\citep{lee2021creators, valle2021transflower, yao2023dance, liang2024intergen}, assisting artists in crafting immersive experiences for both dancers and audiences. To engage the audience, dance must align with the rhythm and emotional tone of the music. Compared to single dance, group dance offers richer visual impact through spatial coordination, synchronized actions, and dynamic formation changes, making choreography significantly more complex. While existing works~\citep{le2023controllable, dai2025harmonious, yang2024codancers} and datasets~\citep{tsuchida2019aist, aiozGdance, zhang2025motion} have advanced the field, two major challenges remain for group dance generation.

\textbf{Generation of long sequences.}
The goal of music-to-dance generation is to synthesize coherent motion sequences aligned with an entire musical composition. Short sequences often fail to capture structural progression and emotional dynamics, such as the buildup before a beat drop. To improve scalability, several works adopt hierarchical or staged strategies that first generate high-level motion structures and then refine them into full sequences~\citep{li2024lodge}. Lodge~\citep{li2024lodge} proposes a two-stage framework for solo dance synthesis by predicting key poses and interpolating intermediate frames, while Lodge++~\citep{li2024lodgepp} extends this idea through a coarse-to-fine choreography design, generating global dance primitives and refining them via diffusion. Although such staged designs ease long-horizon modeling, they rely on stage-wise decomposition, which may accumulate errors and introduce inconsistencies between global structure and local motion. Moreover, these methods focus on single-dancer choreography and do not explicitly address multi-dancer spatial coordination.

For group dance generation, existing methods often produce partially overlapping motion segments and apply smoothing during stitching to maintain continuity~\citep{tseng2023edge}. Diffusion-based models such as GCD~\citep{le2023controllable} and TCDiff~\citep{dai2025harmonious} improve segment consistency during sampling. However, stochastic diffusion sampling may assign inconsistent spatial identities across segments, leading to abrupt positional shifts when merged~\citep{dai2026tcdiff++}. An alternative is to generate the entire sequence in a single pass. Yet long-range sequence modeling is inherently challenging due to increasing temporal dependencies and computational cost~\citep{zhou2021informer, bulatov2024beyond}. In multi-dancer choreography, complexity grows with both sequence length and dancer count. Most end-to-end approaches rely on diffusion transformers, resulting in quadratic complexity ($N^2 \times L^2$). Efficient long-duration group dance generation therefore remains an open challenge.

\textbf{Multi-dancer interaction modeling.}
As previously mentioned, the quality of group dance performances depends on precise inter-dancer coordination. Most existing methods~\citep{le2023controllable,tseng2023edge,yang2024codancers} represent group sequences by concatenating body-motion features and 3D trajectories of all dancers into a unified vector, making it difficult to explicitly model spatial correlations across performers.

GCD~\citep{le2023controllable} employs cross-attention to capture inter-dancer relationships, but this mechanism models only coarse spatial–temporal interactions and may lead to \textbf{overly uniform} motion patterns. TCDiff~\citep{dai2025harmonious} addresses dancer ambiguity through trajectory-controllable diffusion, yet trajectory generation still lacks explicit spatial correlation modeling, which may result in \textbf{motion overlap or collisions}. Its additional dance-beat navigator also increases computational cost. TCDiff++~\citep{dai2026tcdiff++} extends this line of work with an end-to-end trajectory-controllable diffusion framework that improves stability for long-duration choreography. By incorporating positioning embeddings and swap-mode encoding, it enhances global temporal coherence. However, it still relies on transformer-based diffusion modules for spatial–temporal modeling, leading to substantial computational overhead and increased inference latency as sequence length grows. Explicit user-level control over multi-dancer spatial coordination also remains limited.

\begin{figure*}[tb]
    \centering
    \includegraphics[width=0.9\linewidth]{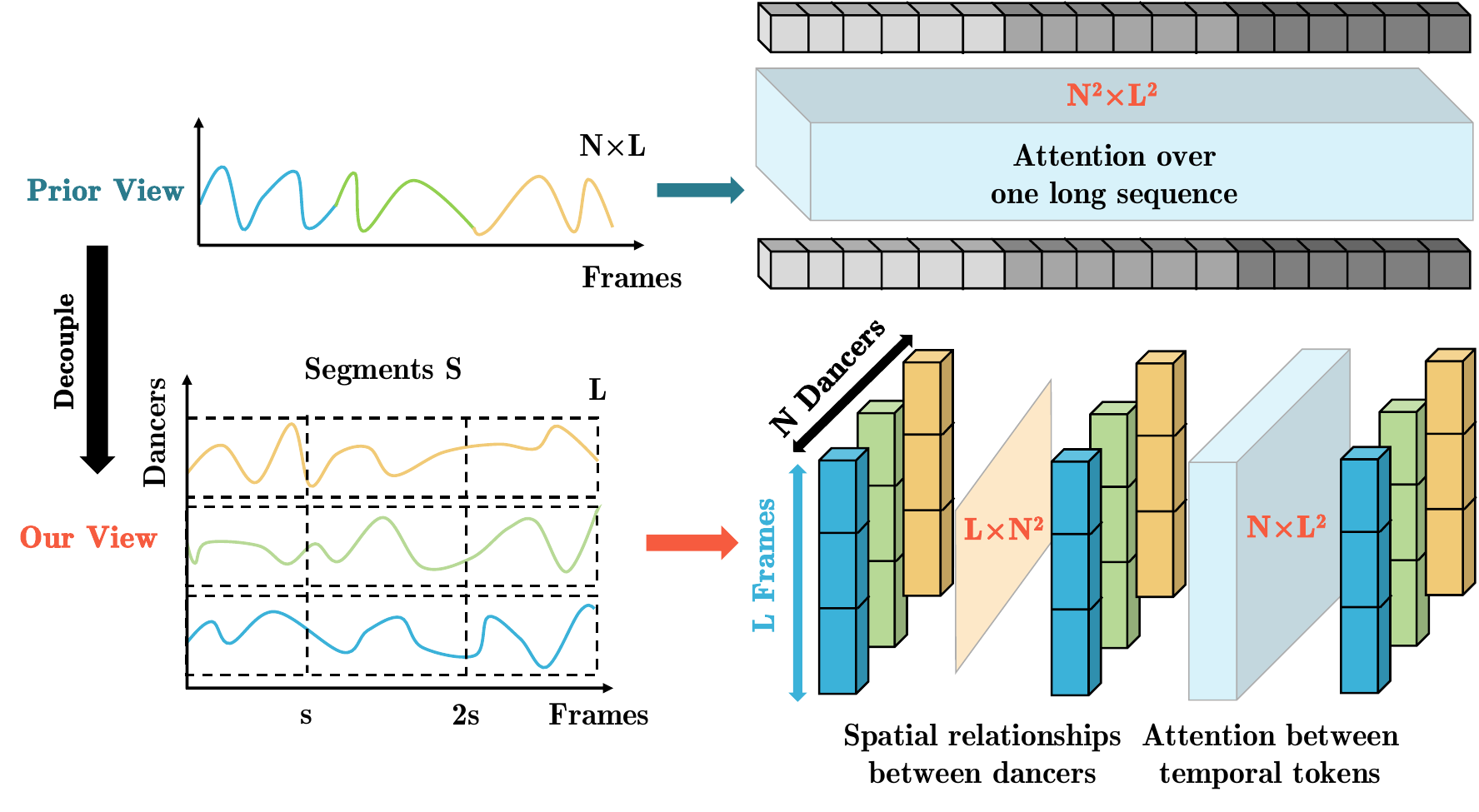}
    \caption{Comparison between vanilla Group Dance Transformer (top) and ours. We decouple the long-duration group dance sequence $N \times L$ into spatial and temporal parts to reduce the computational complexity to $O(LN^2)+O(NL^2)$. The overall complexity can be reduced into near $O(NL)$ (see Section~\ref{tt}).  Also in our framework, the temporal dimension is divided into segments and the noise schedule in training will also change to capture local dependencies and support streaming generation (will be discussed in Section~\ref{tns}).}
\label{fig:design}
\end{figure*}

Our previous conference version published in BMVC 2025, ST-GDance~\citep{xu2025stgdance}, addresses this issue through graph-based spatial modeling to capture inter-dancer relationships. By combining graph neural networks with a lightweight temporal attention module, it reduces computational complexity from $O((N\times L)^2)$ to nearly $O(NL)$ while producing coherent short sequences. 

However, although global coordination is preserved, fine-grained motion details—especially subtle footwork and rapid transitions—are less accurate. This limitation may stem from the use of linear global attention, which can smooth local motion dynamics and remove useful details, resulting in occasional unnatural or jittery motions in longer sequences. This issue negatively affects generation quality. Additionally, ST-GDance may struggle in interactive applications due to its bidirectional attention dependencies, which introduce first-token latency under a non-streaming generation paradigm. In summary, to improve interactivity in real-world applications, ST-GDance needs to be adapted into a streaming framework with enhanced motion fidelity.

To address these limitations, we propose \textbf{ST-GDance++}, a scalable and streaming spatial–temporal diffusion framework for long-duration group dance generation. First, inspired by motion generation work~\citep{yin2025slow}, we introduce a segment-based triangular noise schedule (TNS) that enables progressive denoising over long sequences, improving motion continuity and robustness during streaming generation. Second, to balance efficiency and motion fidelity, we design an alignment attention mask (AAM) inspired by sparse dependency modeling~\citep{cai2025flooddiffusion} to focus on locally relevant conditioning signals. An efficient state space model (SSM) is further employed to aggregate global temporal dependencies. Together with TNS, this design stabilizes long-term generation while reducing uncertainty accumulation. Third, instead of dense attention over $N \times L$ tokens, which incurs quadratic complexity and ignores explicit spatial coordination, we retain the graph-based spatial modeling block (SMB) from ST-GDance. It encode distance-aware inter-dancer relationships, enabling accurate spatial interaction with low computational overhead. Overall, ST-GDance++ generates long, synchronized group dance sequences in a streaming manner while preserving precise inter-dancer coordination and temporal coherence.

Our contributions can be summarized as follows:

\begin{itemize}
    \item \textbf{Segment-Based Triangular Noise Schedule (TNS).} We reformulate the noise schedule by introducing segment-wise correlated noise during training. This strategy significantly enhances model robustness and enables seamless streaming inference, thereby improving the overall practical utility and user experience in real-world applications.
    
    \item \textbf{Efficient Temporally Aligned Attention Mask (AAM).} We introduce a masked cross-attention mechanism that restricts each frame to locally relevant conditioning signals, improving motion–condition alignment while reducing redundant dependencies. This design lowers attention complexity from quadratic to linear time without sacrificing bidirectional information flow.

    \item \textbf{Scalable Spatial–Temporal Framework for Group Dance Modeling.} We introduce a novel dance decoder that innovatively decouples spatial coordination from temporal dynamics (Figure~\ref{fig:design}). This decomposition transforms quadratic token interactions into an additive spatial–temporal complexity, enabling both computationally efficient and globally coherent long-duration motion synthesis.
    
    \item \textbf{State-of-the-Art Group Choreography Quality.} Extensive experiments on AIOZ-GDance dataset demonstrate that ST-GDance++ scales to longer sequences while achieving superior motion quality and significantly improved computational efficiency compared with existing group choreography methods.

\end{itemize}

\section{Related Work}\label{related}

\subsection{Music-driven Single-Dancer Generation}
Music-to-dance generation aims to synthesize motion sequences that are both natural and temporally coherent with the input music~\citep{joshi2021extensive,sui2026survey}. Early studies mainly relied on rule-based or motion-graph techniques~\citep{kovar2002motion,kim2003rhythmic,safonova2007construction}, while recent work has increasingly adopted deep learning approaches, especially generative models~\citep{zhu2023human,yin2023dance,sun2020deepdance,ferreira2021learning}. Most existing methods take music features together with previously generated motion as inputs. Although this design enables sequential prediction, it often leads to error accumulation over time and limits motion diversity. In addition, many approaches struggle to capture long-range temporal dependencies, which reduces the coherence and novelty of generated sequences.

With the emergence of large-scale dance datasets~\citep{lee2019dancing,li2021ai,li2023finedance}, generative modeling has become the dominant paradigm. Auto-regressive methods~\citep{alemi2017groovenet,yalta2019weakly} generate motion step by step, but are prone to drift over long sequences. To improve diversity and realism, multimodal generative frameworks based on GANs~\citep{lee2019dancing}, VAEs~\citep{hong2022avatarclip,siyao2022bailando,siyao2023bailando++}, and diffusion models~\citep{tseng2023edge} have been proposed. Despite these advances, two major challenges remain: (1) most models are trained on short motion clips, making it difficult to generalize to long-duration sequences; and (2) methods designed for single dancers often fail to handle group scenarios, where both temporal synchronization and spatial coordination are required.

\subsection{Music-driven Multi-Dancer Generation}

Compared with single-dancer motion synthesis, music-driven multi-dancer generation remains relatively underexplored. Only a limited number of works~\citep{yao2023dance,le2024scalable,le2023controllable,yang2024codancers,dai2025harmonious} consider scenarios involving more than two dancers, where both temporal synchronization and spatial coordination are essential.

Early approaches such as GDanceR~\citep{aiozGdance} and GCD~\citep{le2023controllable} extend single-dancer models to group settings, but they do not explicitly address motion representation imbalance across dancers, often resulting in ambiguous formations. CoDancers~\citep{yang2024codancers} decomposes group choreography into individual motion streams and progressively introduces dancers, reducing computational cost but potentially overlooking global formation consistency. TCDiff~\citep{dai2025harmonious} proposes a trajectory-guided framework that first predicts dancer coordinates and then generates motions conditioned on these trajectories, improving spatial consistency and reducing collisions. However, its two-stage pipeline separates trajectory prediction from motion synthesis, which may introduce inconsistencies in longer sequences. TCDiff++~\citep{dai2026tcdiff++} reformulates this design into an end-to-end trajectory-controllable diffusion framework, enhancing stability and temporal coherence for long-duration choreography. Nevertheless, it still relies on transformer-based diffusion modules with bidirectional attention, leading to quadratic complexity with respect to sequence length and dancer count.

Our previous work, ST-GDance~\citep{xu2025stgdance}, adopts a spatial–temporal graph formulation to model inter-dancer interactions and motion dynamics jointly. While it improves coordination compared with earlier methods, its scalability for long-duration generation remains limited. In this work, we extend it to ST-GDance++, a scalable spatial–temporal diffusion framework that explicitly models spatial interactions while enabling efficient long-sequence generation, achieving improved coordination and scalability in multi-dancer scenarios.

\subsection{Streaming Generation with Diffusion Model}

Streaming generation produces sequences progressively, allowing the model to adjust outputs when new control signals arrive instead of synthesizing the entire sequence at once~\citep{zhang2025primal,kodaira2025streamdit,jiang2025causal}. This enables faster feedback and more practical interactive creation.

Recent studies incorporate streaming into diffusion models via diffusion forcing, assigning different noise levels to frames or tokens for flexible denoising~\citep{benoit2025diffusion}. Subsequent work mitigates train–test mismatch through explicit rollout strategies such as self forcing and rolling forcing~\citep{huang2025self,liu2025rolling}. Other methods enforce causal or monotonic timestep constraints to preserve temporal consistency~\citep{chen2025skyreels,sun2025ar}. However, most approaches are designed for video generation with spatially large 2D inputs. Motion synthesis, by contrast, is temporally one-dimensional and often driven by rapidly changing controls, making direct adaptation suboptimal.

Streaming diffusion has recently been extended to motion generation. FloodDiffusion~\citep{cai2025flooddiffusion} proposes a tailored diffusion-forcing strategy for text-driven streaming motion synthesis. Yet existing methods primarily address single-agent settings. Group dance generation introduces additional challenges, including spatial coordination and synchronized transitions among multiple dancers. Our work adapts streaming diffusion to multi-dancer choreography, enabling long-duration generation while maintaining coherent spatial structures.

\section{Preliminary Knowledge}\label{back}

\subsection{Problem Definition}\label{pro}

Given an input music sequence $\mathcal{M} = \{ m^l \}_{l=1}^{L}$, the objective of group dance generation is to synthesize a corresponding multi-dancer motion sequence $\mathcal{X} = \{ x^l \}_{l=1}^{L}$, where $l$ denotes the frame index and $L$ is the sequence length. Each frame $x^l$ contains the poses of all dancers, defined as $x^l = \left\{ x^{l,n} \right\}_{n=1}^{N},$ where $x^{l,n}$ represents the pose of the $n$-th dancer at frame $l$, and $N$ is the total number of dancers. Accordingly, the motion sequence of the $n$-th dancer can be written as $\mathcal{X}^n = \{ x^{l,n} \}_{l=1}^{L}$. Each dancer pose is represented by a 151-dimensional vector, including 24-joint SMPL~\citep{loper2023smpl} with 6D rotations~\citep{zhou2019continuity}, binary contact indicators, and 3D root position~\citep{mcfee2015librosa}. Therefore, the group motion sequence can be represented as a tensor $\mathcal{X} \in \mathbb{R}^{L \times N \times 151}$.
Similarly, the music sequence is encoded into frame-level feature vectors using the audio encoder, resulting in a music feature tensor $\mathcal{M} \in \mathbb{R}^{L \times d_m}$ where $d_m$ denotes the dimensionality of the music feature from Librosa~\citep{mcfee2015librosa}. Under the diffusion formulation, we denote $x_t$ as the noisy motion at diffusion timestep $t$, and the model is trained to progressively denoise $x_t$ toward the clean motion sequence conditioned on the music input $\mathcal{M}$.

\subsection{Diffusion for Group Dance Generation}\label{diffusion}

Diffusion models have demonstrated strong capabilities in high-dimensional generative tasks such as images~\citep{saharia2022photorealistic}, audio~\citep{guo2024audio}, and videos~\citep{qi2023diffdance}. Following the denoising diffusion probabilistic model (DDPM)~\citep{ho2020denoising} and its application to dance generations~\citep{tseng2023edge,le2023controllable,dai2025harmonious,dai2026tcdiff++}, we adopt a conditional diffusion framework for group motion synthesis.

The diffusion process constructs a Markov chain that gradually perturbs the clean motion sequence $x_0$ into a Gaussian noise distribution through $T$ steps. The forward process is defined as:
\begin{equation}
q(x_t | x_{t-1}) = \mathcal{N}(x_t; \sqrt{1 - \beta_t}x_{t-1}, \beta_tI),
\end{equation}
where $x_t$ denotes the noisy sample at timestep $t$, and $\beta_t$ controls the variance of the injected noise. During generation, the reverse process aims to recover the clean motion from noise. The reverse transition is modeled as:
\begin{equation}
p(x_{t-1} | x_t) = \mathcal{N}(x_{t-1}; \frac{1}{\sqrt{1 - \beta_t}}(x_t - \beta_t\mu_t), \sigma_t^2 I),
\end{equation}
where $\mu_t$ is the model-predicted mean and $\sigma_t^2$ denotes the variance of the reverse process.

To generate group dance sequences conditioned on music and spatial context, we extend the diffusion model into a conditional formulation. Given the music condition $\mathcal{M}$ and the spatial condition $\mathcal{S}$, the denoising network $D$ learns to reverse the diffusion trajectory:
\begin{equation}
x_{t-1} = D(x_t, t, \mathcal{M}, \mathcal{S}),
\end{equation}
where the network predicts the clean motion or its equivalent parameterization at each timestep. In this way, the model progressively transforms random noise into a coherent multi-dancer motion sequence aligned with the input music.

\subsection{State Space Model (SSM)}\label{ssm}

State Space Models (SSMs) have recently shown strong capability in modeling long-range sequential data due to their linear-time complexity and effective global context aggregation~\citep{gu2022ssm,weng2024mamballie}. Compared with conventional attention-based architectures, SSMs provide a more scalable solution for long-duration motion generation due to its superior feature selection ability.

Given an input sequence $x(t) \in \mathbb{R}$, an SSM transforms it into an output $y(t) \in \mathbb{R}$ through a latent state $h(t) \in \mathbb{R}^{N}$ governed by a linear dynamical system:

\begin{equation}
h'(t) = Ah(t) + Bx(t), \quad y(t) = Ch(t),
\end{equation}

\noindent where $A \in \mathbb{R}^{N \times N}$ controls the state transition, and $B \in \mathbb{R}^{N \times 1}$ and $C \in \mathbb{R}^{1 \times N}$ denote the input and output projection matrices, respectively.

To enable efficient computation, we adopt the Mamba-style discretization with zero-order hold (ZOH)~\citep{gu2024mamba}. The continuous system is converted into a discrete form:

\begin{equation}
\bar{A} = \exp(\Delta A), \quad
\bar{B} = (\Delta A)^{-1} (\exp(\Delta A) - I)\Delta B,
\end{equation}

\noindent where $\Delta$ is an input-dependent step size that determines the temporal resolution. Smaller $\Delta$ focuses on short-term variations, while larger $\Delta$ captures longer-range dependencies. The discretized system can be expressed as a convolution with a structured kernel:

\begin{equation}
\bar{K} = (\bar{C}\bar{B}, \bar{C}\bar{A}\bar{B}, \dots, \bar{C}\bar{A}^{L-1}\bar{B}), 
\quad
y = x * \bar{K},
\end{equation}

\noindent where $L$ denotes the sequence length. This formulation enables the model to efficiently aggregate information across the entire temporal horizon. In our framework, the SSM backbone serves as the temporal modeling module, capturing long-range motion dynamics while maintaining linear complexity with respect to sequence length $O(L)$. This property is particularly beneficial for long-duration group dance generation. 

\section{Methodology}\label{method}

\begin{figure*}[tb]
    \centering
    \includegraphics[width=0.9\linewidth]{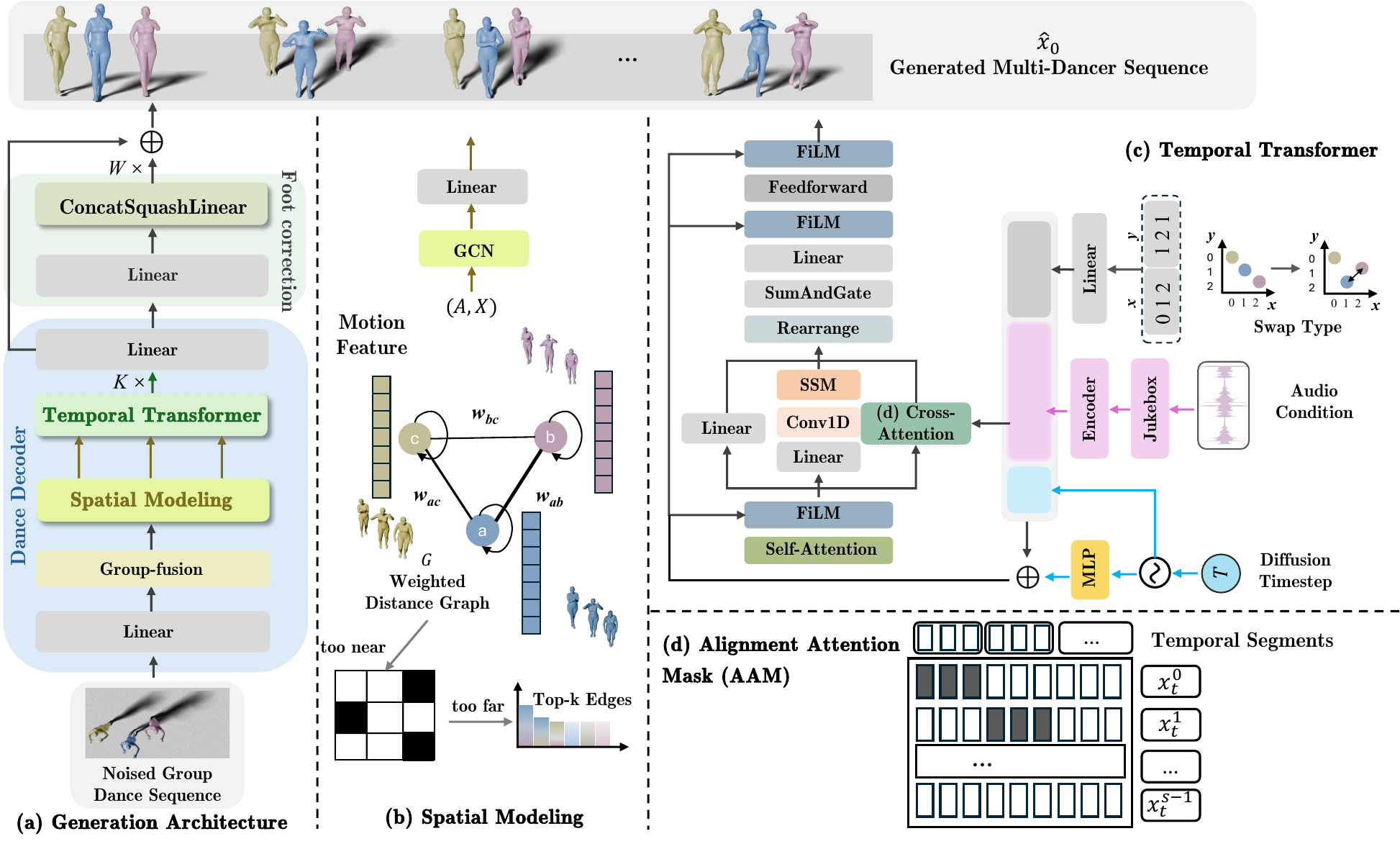}
    \caption{Overview of the proposed ST-GDance++ framework for efficient long-duration group dance generation. (a) Overall generation architecture, where noised group motion is processed by the spatial–temporal diffusion decoder to produce the final multi-dancer sequence. (b) Spatial distance-aware graph modeling, which constructs a weighted graph based on pairwise dancer distances and applies a lightweight GCN to capture inter-dancer interactions. (c) Temporal transformer module, responsible for modeling long-range motion dynamics under music and timestep conditions. (d) Alignment attention mask (AAM), which enforces temporally consistent interactions across segments to support stable streaming generation.}
\label{fig:framework}
\end{figure*}

In this section, we introduce ST-GDance++, a spatial–temporal diffusion framework for scalable and high-precision generation of long-duration group dance sequences, as illustrated in Fig.~\ref{fig:framework}. Unlike conventional approaches that treat dancer–frame tokens as a single sequence, we explicitly decouple the dancer and temporal dimensions, modeling spatial interactions and temporal dynamics with dedicated modules. The \textbf{spatial modeling block (SMB)} employs a distance-aware graph convolutional network to capture inter-dancer relationships through distance-aware graph interactions. The temporal branch focuses on motion evolution over time. To support stable long-duration generation, we further introduce an \textbf{Alignment Attention Mask (AAM)} and a \textbf{Segment-based Triangular Noise Schedule (TNS)}, which improve temporal consistency and robustness during diffusion. 

\subsection{Group Fusion Projection}
To reduce dancer ambiguity and encourage group-level interaction, we use a lightweight group fusion module following previous work~\citep{dai2025harmonious}. Instead of processing each dancer independently, the features of all dancers are first concatenated along the channel dimension to form a shared representation because the data in high-dimensional feature spaces can be more easily differentiated. Specifically, the input group feature tensor $X \in \mathbb{R}^{L \times N \times d}$, representing $N$ dancers with sequence length $L$ and feature dimension $d$, is first reshaped into  $X' \in \mathbb{R}^{L \times (N \times d)}$. This fused feature is then projected with a multi-layer perceptron (MLP) to produce higher-dimensional latent space feature for all dancers. Finally, the features are reshaped back into dancer-specific representations for subsequent spatial–temporal modeling.


\subsection{Spatial-temporal Group Dance Decoder}
As shown in Fig~\ref{fig:design}, we decouple multi-dancer generation task into spatial and temporal dynamics and use different blocks in designed dance decoder to model separately in Fig.~\ref{fig:framework}.
\subsubsection{Spatial Distance-aware Graph Convolutional Network} 
In group dance generation, each frame contains two essential factors: the interactions among dancers and the global formation of the group. Many existing approaches primarily focus on temporal motion synthesis, while spatial interactions between dancers are only implicitly modeled or even neglected. This often leads to ambiguous spatial relationships or motion collisions in crowded scenes.

To explicitly capture intra-group spatial dependencies, we represent the dancers at each frame as nodes in a fully connected weighted graph as shown in Fig.~\ref{fig:framework} (b). The edge weights are determined by pairwise Euclidean distances, allowing the model to encode distance-aware interactions. Let $N$ denote the number of dancers. The adjacency matrix $A \in \mathbb{R}^{N \times N}$ is defined as:
\begin{equation}
A_{ij} = \frac{1}{\| p_i - p_j \| + \epsilon},
\end{equation}
where $p_i$ and $p_j$ are the 2D coordinates of dancers $i$ and $j$, and $\epsilon$ is a small constant that prevents numerical instability. This formulation assigns larger weights to closer dancers, encouraging stronger interaction modeling in spatially relevant regions. Given the latent motion feature sequence $X^{group} \in \mathbb{R}^{L \times N \times d}$ after group fusion projection, where $L$ is the sequence length and $d$ is the feature dimension, spatial features are propagated using a graph convolutional layer:
\begin{equation}
H^{(l+1)} = \text{ReLU}\left(\tilde{\mathbf{A}} H^{(l)} W^{(l)}\right),
\end{equation}
where $\tilde{\mathbf{A}} = D^{-1/2} \mathbf{A} D^{-1/2}$ is the normalized adjacency matrix and $W^{(l)}$ is a learnable transformation.

To control computational cost, we apply a top-$k$ edge selection strategy, keeping only the most significant connections for each node. This reduces the per-layer complexity to:
\begin{equation}
O_{\text{GCN}} = O(L N k) \approx O(LN),
\end{equation}
where $k$ is a fixed proportion of retained edges. In addition, extremely close connections are masked to prevent overly strong correlations that may lead to spatial collisions.

By introducing this lightweight distance-aware GCN, spatial interactions are modeled explicitly without relying on dense cross-attention over all dancer–frame tokens. This design reduces computational redundancy, improves scalability, and enables more stable and collision-free group motion generation.

\subsubsection{Temporal Transformer}\label{tt}

The temporal module models long-range motion dynamics along the time dimension, as illustrated in Fig.~\ref{fig:framework}(c).

\textbf{Self-attention.}  
To capture long-range temporal dependencies, we adopt 
Differential Attention from the DIFF Transformer~\citep{ye2025differential} 
as the temporal self-attention mechanism. 
Unlike standard multi-head attention, this design computes 
attention as the difference between two parallel attention branches, 
which enhances discriminative information filtering. 
The module is illustrated in the temporal decoder 
(Fig.~\ref{fig:framework}(c)).

Given a single dancer feature sequence 
$X \in \mathbb{R}^{L \times d}$ from spatially processed group feature $X^{GCN} \in \mathbb{R}^{L \times N \times d}$,
we obtain two sets of query–key pairs by splitting 
the channel dimension of linear projections:

\begin{equation}
[Q_1; Q_2] = XW_Q,  [K_1; K_2] = XW_K,  V = XW_V,
\end{equation}

\noindent where $XW_Q, XW_K \in \mathbb{R}^{L \times  2d}$, 
and $Q_1, Q_2, K_1, K_2 \in \mathbb{R}^{L \times d}$ 
are obtained by evenly dividing the last dimension, 
while $V \in \mathbb{R}^{L \times 2d}$. The two attention maps are computed as

\begin{equation}
A_1 = \text{softmax}\!\left( \frac{Q_1 K_1^\top}{\sqrt{d}} \right), 
A_2 = \text{softmax}\!\left( \frac{Q_2 K_2^\top}{\sqrt{d}} \right),
\end{equation}

\noindent and the differential attention output is defined as

\begin{equation}
\text{DiffAttn}(X) = (A_1 - \lambda A_2)V,
\end{equation}

\noindent where $\lambda$ is a learnable suppression coefficient 
that balances the two branches.

Because the projections for the two branches share the same 
linear transformations, the projection cost remains 
$O(Ld^2)$ and does not dominate computation. Forming the attention matrices 
$Q_1K_1^\top$ and $Q_2K_2^\top$ 
requires $O(L^2d)$ operations in theory. 
However, the difference matrix $A_1 - \lambda A_2$ 
is empirically sparse—most entries are close to zero and 
are pruned during implementation—so each query retains 
at most $s$ significant responses with $s \ll L$.

As a result, the effective number of query–key interactions 
reduces from $L^2$ to $Ls$, leading to a per-layer 
complexity of $O(Lsd)$, which further simplifies to 
$O(Ld)$ when $s$ is treated as a constant. 
Therefore, the temporal modeling scales linearly 
with respect to the sequence length.

\textbf{Cross-attention with alignment mask.}  
To incorporate music conditions, we apply cross-attention between motion tokens and music features, as shown in Fig.~\ref{fig:framework}(d). A key challenge in long-duration and streaming generation is that dense cross-attention allows each motion frame to attend to the entire music sequence, which may introduce temporal inconsistency and cause abrupt motion changes when the sequence is processed segment by segment. To address this issue, we introduce an Alignment Attention Mask (AAM) that restricts each frame to attend only to a temporally aligned local music window, enforcing consistent music–motion correspondence across segments.

Let $M \in \mathbb{R}^{L \times d}$ denote the music features. 
The cross-attention is formulated as:
\begin{equation}
\text{CA}(X, M) =
\text{Softmax}\!\left(
\frac{QK_m^\top + A}{\sqrt{d}}
\right)V_m,
\end{equation}
\noindent where
$Q = X W_Q^m \in \mathbb{R}^{L \times d}$,
$K_m = M W_K^m \in \mathbb{R}^{L \times d}$,
$V_m = M W_V^m \in \mathbb{R}^{L \times d}$,
and $A \in \mathbb{R}^{L \times L}$ is the Alignment Attention Mask (AAM).
This design provides two advantages. 
First, it enforces temporally consistent music–motion alignment, reducing abrupt transitions when generating long sequences. 
Second, it significantly reduces computational cost. 
Assuming each frame attends only to a local window of size $w$, the complexity becomes:
\begin{equation}
O_{\text{cross}} = O(Lwd),
\end{equation}
which scales linearly with the sequence length when $w$ is fixed and $w \ll L$.

\textbf{SSM-enhanced temporal modeling.}  
To efficiently aggregate global context without re-introducing quadratic overhead, the features are subsequently processed by an SSM block (introduced in Sec.~\ref{ssm}) acting as a lightweight temporal backbone:
\begin{equation}
X^{SSM} = \text{SSM}(X),
\end{equation}
with complexity linear in the sequence length:
\begin{equation}
O_{\text{SSM}} = O(Ld).
\end{equation}

\textbf{Swap mode and rearrangement.}
Following TCDiff++\citep{dai2026tcdiff++}, we explicitly model dancer permutation within each temporal segment via a swap-mode condition $\mathcal{S}$, and use a canonical rearrangement to stabilize dancer identities across samples.
Given the clean group motion $\mathcal{X}=\{x^l\}_{l=1}^{L}$, we first \emph{rearrange} dancers in the initial frame by sorting them according to their horizontal positions (x-axis), so that dancer indices are consistent across the dataset (i.e., dancers are ordered from left to right in the first frame). 
The diffusion process is then applied to the rearranged motion to obtain the noisy input $x_t$.

To describe whether dancers exchange their relative spatial orders within a segment, we compute the swap type by tracking the ordering changes between the start and end frames of each segment.
Concretely, we assign dancer IDs according to their initial spatial ranks (left-to-right and top-to-bottom, as illustrated in Fig.~\ref{fig:framework}(c)). 
At the end of the segment, we re-rank dancers along the x- and y-axes and record the resulting index orders, which are concatenated into an index sequence. 
This index sequence captures the permutation pattern within the segment and is further embedded into a high-dimensional vector $\mathcal{S}\in\mathbb{R}^{d}$, which is fused with other conditions (e.g., music features and diffusion timestep) to guide denoising.

\textbf{Overall complexity.}  
Combining self-attention, cross-attention, and the SSM block, the overall temporal complexity is:
\begin{equation}
O_{\text{temporal}} 
= O(Lsd) + O(Lwd) + O(Ld) 
\approx O(Ld),
\end{equation}
which scales linearly with the sequence length and enables efficient long-duration group dance generation.

\subsection{Segment-based Triangular Noise Schedule (TNS)} \label{tns}
\begin{figure}[tb]
    \centering
    \includegraphics[width=\linewidth]{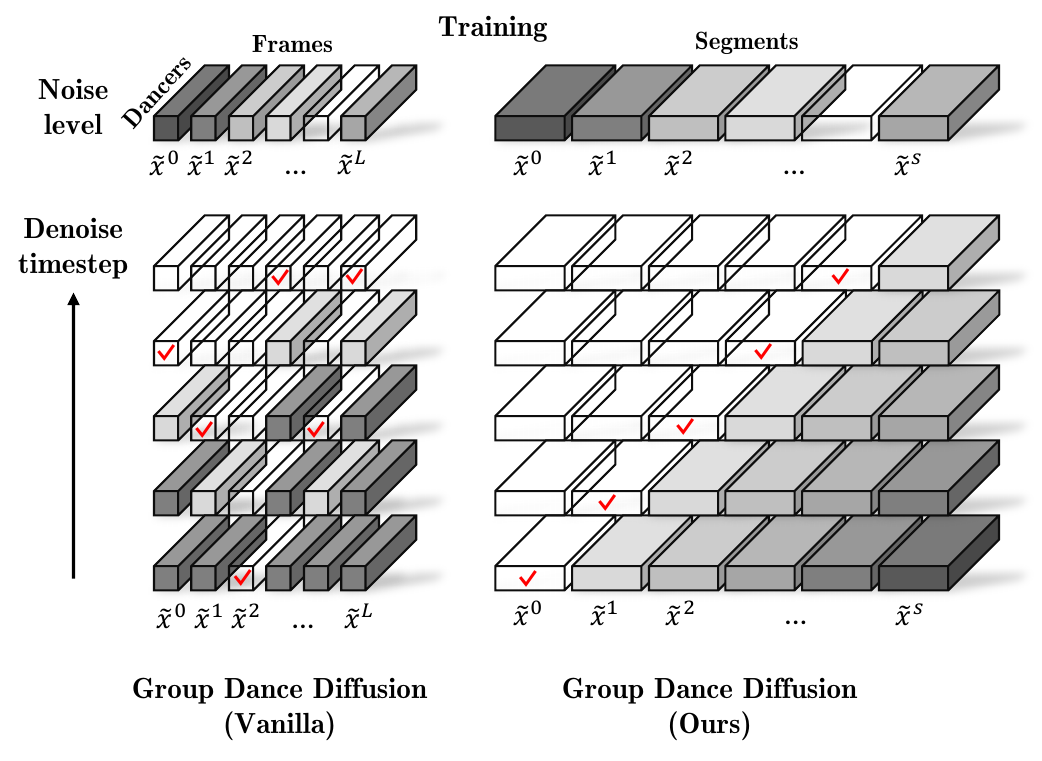}
    \caption{Noise schedule comparison. Diffusion Forcing typically samples training schedules with stochastic active windows, leading to a noticeable discrepancy between training and inference. To bridge this gap, we introduce a segment-based triangular schedule that exclusively denoises the current active window before advancing. By applying a consistent noise level to all frames within a single segment, our approach ensures temporal coherence and enhances the stability of long-duration generation.}
\label{fig:seg-diffusion}
\end{figure}

To bridge the discrepancy between training and inference phases, we reformulate the diffusion process using a segment-based triangular noise schedule inspired by~\citep{cai2025flooddiffusion}, as shown in Fig.~\ref{fig:seg-diffusion}. Unlike vanilla Diffusion Forcing that assigns stochastic noise levels to individual frames, our approach ensures temporal coherence by processing the sequence in structured segments.

We divide the length $L$ group dance sequence $x$ into $s$ segments, where each segment $x^s$ consists of a fixed number of frames. During training, for a sampled global timestep $t \in [1, T]$, the noised segment $\tilde{x}^s$ is formulated as:
\begin{equation}
\tilde{x}^s = \sqrt{\bar{\alpha}_t} x^s + \sqrt{1 - \bar{\alpha}_t} \epsilon, \quad \epsilon \sim \mathcal{N}(0, \mathbf{I})
\end{equation}
Crucially, all frames within the segment $x^s$ share a consistent noise level $\sqrt{\bar{\alpha}_t}$. As illustrated in the triangular schedule, the model is trained to denoise the current active window while conditioned on the historical context $x^{<s}$. This design offers two key advantages for long-term stability: (1) By applying a uniform noise level within each segment $x^s$, the model maintains \textbf{temporal invariance} and avoids learning spurious frame-to-frame noise transitions, which effectively enhances local motion smoothness; (2) The triangular progression provides \textbf{error robustness} by explicitly simulating the auto-regressive inference process, thereby training the model to predict coherent subsequent segments from varying historical noise levels and significantly mitigating the accumulation of drifting errors during extended generation.

\subsection{Training Losses}

Our ST-GDance++ is trained in an end-to-end manner under a conditional diffusion framework. Following previous works~\citep{tseng2023edge,dai2026tcdiff++}, the objective combines the standard diffusion reconstruction loss with several motion-level constraints to ensure temporal coherence, physical plausibility, and stable group interactions.

\textbf{Diffusion Reconstruction loss.}
We adopt the standard diffusion loss to encourage the denoised motion to match the ground-truth sequence:
\begin{equation}
L_{\text{simple}} = \mathbb{E}_{x,t} 
\left\| x_0 - D(x_t, t, \mathcal{M}, \mathcal{S}) \right\|_2^2 ,
\end{equation}
where $x_0$ denotes the clean motion, $x_t$ is the noised motion at timestep $t$, $M$ is the music condition, and $D(\cdot)$ is the denoising network.

\textbf{Joint Velocity loss.}
To enforce temporal smoothness between consecutive frames, we introduce a velocity loss:
\begin{equation}
L_{\text{vel}} = 
\frac{1}{L-1} \sum_{l=1}^{L-1}
\left\|
(x_0^{l+1} - x_0^{l}) -
(\hat{x}_0^{l+1} - \hat{x}_0^{l})
\right\|_2^2 .
\end{equation}

\textbf{Forward Kinematics loss.}
To preserve the spatial consistency of the skeletal structure, we apply a forward kinematics constraint:
\begin{equation}
L_{\text{FK}} =
\frac{1}{L} \sum_{l=1}^{L}
\left\|
FK(x_0^{l}) - FK(\hat{x}_0^{l})
\right\|_2^2 ,
\end{equation}
where $FK(\cdot)$ denotes the forward kinematics function.

\textbf{Foot Contact loss.}
To improve physical realism and reduce foot sliding, we employ a foot contact constraint:
\begin{equation}
L_{\text{contact}} =
\frac{1}{L-1} \sum_{l=1}^{L-1}
\left\|
\left(FK(\hat{x}_0^{l+1}) - FK(\hat{x}_0^{l})\right)
\cdot \hat{d}^{l}
\right\|_2^2 ,
\end{equation}
where $\hat{d}^{l}$ indicates the predicted contact state.

\textbf{Distance consistency loss.}
To preserve stable spatial relationships among dancers, \cite{dai2026tcdiff++} introduce a distance-consistency loss that penalizes deviations in pairwise distances between predicted and ground-truth motions:
\begin{equation}
\begin{aligned}
L_D
=&\;
\frac{1}{N-1}
\sum_{l=1}^{L}
\binom{N}{2} \\
&\cdot
\left\|
\left(
p^{(l),i} - p^{(l),j}
\right)
-
\left(
\hat{p}^{(l),i} - \hat{p}^{(l),j}
\right)
\right\|_2^2 .
\end{aligned}
\end{equation}
where $p^{(l),i}$ and $\hat{p}^{(l),i}$ denote the ground-truth and predicted positions of dancer $i$ at frame $l$, respectively. 

This loss encourages the model to maintain consistent inter-dancer distances, reducing collisions and preserving group formations.

\textbf{Overall objective.}
The final training loss is a weighted combination of the above terms:
\begin{equation}
\begin{aligned}
Loss =\;&
\lambda_{\text{simple}} L_{\text{simple}} +
\lambda_{\text{FK}} L_{\text{FK}} +
\lambda_{\text{vel}} L_{\text{vel}} \\
&+
\lambda_{\text{contact}} L_{\text{contact}} +
\lambda_{\text{D}} L_{\text{D}} .
\end{aligned}
\end{equation}
where $\lambda_{\text{simple}}$, $\lambda_{\text{FK}}$, $\lambda_{\text{vel}}$, $\lambda_{\text{contact}}$, and $\lambda_{\text{D}}$ are balancing coefficients.

\section{Experiments}\label{experiments}
\subsection{Experimental Settings}

\textbf{Implementation Details.} All experiments are implemented in PyTorch and conducted on NVIDIA RTX 4090 and V100 GPUs. For fair comparison, all efficiency evaluations are performed on  RTX 4090. 
Following previous work~\citep{dai2026tcdiff++}, we set the loss weights as $\lambda_{\text{simple}}=0.636$, $\lambda_{\text{vel}}=2.964$, $\lambda_{\text{FK}}=0.646$, $\lambda_{\text{D}}=100$, and $\lambda_{\text{contact}}=10.942$.
The batch sizes are 64, 32, 24, and 8 per GPU for 2, 3, 4, and 5 dancers, respectively.
We optimize the model using the Adam optimizer with a learning rate of $5\times10^{-5}$.

\begin{table*}[tb]
\centering
\caption{Performance and efficiency comparison on the short-duration group dance setting. $\downarrow$ indicates lower is better, $\uparrow$ indicates higher is better. \textbf{Best} and \underline{runner-up} scores are highlighted.}

\footnotesize
\setlength{\tabcolsep}{3.5pt}

\begin{tabular}{lcccccccccc}
\toprule

\multirow{2}{*}{Method}
& \multicolumn{3}{c}{Group Dance}
& \multicolumn{3}{c}{Single Dance}
& \multicolumn{4}{c}{Efficiency} \\

\cmidrule(lr){2-4}
\cmidrule(lr){5-7}
\cmidrule(lr){8-11}

& GMR$\downarrow$
& GMC$\uparrow$
& TIF$\downarrow$
& FID$\downarrow$
& Div$\uparrow$
& PFC$\downarrow$
& \makecell{FLOPs\\(G)}
& \makecell{Params\\(M)}
& \makecell{Train\\(min/epoch)}
& \makecell{Inf\\(s)} \\

\midrule

EDGE & 70.26 & 62.24 & 0.42 & 33.24 & 10.11 & 3.07 & 12.50 & 52.13 & 0:24 & 8.6 \\
GCD & 30.22 & 80.22 & 0.19 & 39.24 & 9.64 & 2.53 & 27.69 & 62.16 & 1:04 & 4.8 \\
CoDancers & 26.34 & 74.22 & \textbf{0.10} & \underline{23.98} & 9.48 & 3.26 & 53.95 & 56.83 & 0:56 & 4.5 \\
TCDiff++ & \underline{11.85} & \underline{82.04} & \textbf{0.10} & \textbf{22.42} & \underline{17.43} & 1.32 & 32.56 & 70.42 & 1:32 & 9.3 \\
ST-GDance & 14.76 & 81.24 & \underline{0.11} & 28.87 & 12.83 & \textbf{0.97} & \textbf{6.75} & \textbf{50.21} & \underline{0:37} & \textbf{3.3} \\
ST-GDance++ & \textbf{10.85} & \textbf{82.83} & \underline{0.11} & 25.34 & \textbf{18.76} & \underline{1.24} & \underline{11.32} & \underline{51.64} & \textbf{0:42} & \underline{4.2} \\

\bottomrule
\end{tabular}
\label{tab_short}
\end{table*}

\textbf{Dataset.} We conduct experiments on the open-source group dance dataset AIOZ-GDance~\citep{aiozGdance}, which provides 6.7 hours of group dance videos with synchronized music and 3D motion data, covering 4,000 dancers, 7 dance styles, and 16 music genres. All videos are decoded at 30 FPS. Following the protocol of Li et al.~\citep{aiozGdance}, we split the dataset into training, validation, and test sets with a ratio of 8:1:1.

\textbf{Baselines.} We evaluate ST-GDance++ against the following five state-of-the-art group and single dance generation method:
\begin{itemize}
    \item EDGE~\citep{tseng2023edge} is a representative and publicly available single-dancer generation model. For fair comparison, we re-train it on the AIOZ-GDance dataset to evaluate its generalization to group dance settings.
    \item GCD~\citep{le2023controllable} incorporates contrastive learning into a diffusion framework to model dancer–group associations and enables controllable trade-offs between motion consistency and diversity.
    \item CoDancers~\citep{yang2024codancers} is a retrieval-based group dance generation framework that models choreographic units and explicitly decomposes spatial–temporal coherence into individual movements and group interactions.
    \item TCDiff++~\citep{dai2026tcdiff++} adopts a diffusion-based framework with enhanced temporal coherence modeling to generate stable long-duration dance sequences.
    \item ST-GDance~\citep{xu2025stgdance} is our previous work, which adopts a spatial–temporal decoupling strategy with graph-based interaction modeling for efficient and collision-aware group choreography generation.

\end{itemize}

\textbf{Metrics.} For evaluation, we compute both single- and group-level metrics to comprehensively assess the performance of our method: For group evaluation~\citep{le2023controllable}, (1) Group Motion Realism (GMR) quantifies the realism of group formations by computing the Frechet Inception Distance (FID) between generated and real motion features. (2) Group Motion Correlation (GMC) quantifies group coherence by computing the cross-correlation between the motion sequences of generated dancers. (3) Trajectory Intersection Frequency (TIF) quantifies collision frequency by counting the number of trajectory intersections among dancers. For single-dancer generation, (4) Frechet Inception Distance (FID)~\citep{li2021ai, huang2020dance} measures the distribution distance between generated and real motion features. (5) Generation Diversity (Div)~\citep{li2021ai, huang2020dance} quantifies motion diversity based on kinematic features. In addition, to evaluate computational efficiency, we report both device-dependent and device-independent metrics, including floating-point operations (FLOPs), parameters count, training time per epoch, and inference time.

\subsection{Quantitative Comparison} \label{quantitative}

\textbf{Evaluating short-duration scenarios}
Table~\ref{tab_short} presents the quantitative comparison under the short-duration setting with 120 frames. On the group dance task, our method achieves the best performance across two core metrics, yielding 10.85 GMR and 82.83 GMC. Compared with existing baselines, this demonstrates more stable motion reconstruction and better inter-dancer coordination. In particular, relative to TCDiff++, we further reduce GMR from 11.85 to 10.85 and improve GMC from 82.04 to 82.83, while maintaining the same low TIF, indicating fewer abrupt swaps and collisions during generation. Consistent improvements are also observed in single-dancer evaluation. Our model achieves the highest motion diversity, outperforming all baselines by a clear margin, while preserving competitive realism with an FID of 25.34 and a low PFC of 1.24. These results suggest that the generated sequences exhibit both richer dynamics and stable temporal consistency. Overall, our method performs favorably on most group-level coordination metrics and remains competitive on single-dancer quality measures. 

\begin{table*}[tb]
\centering
\caption{Performance and efficiency comparison on the long-duration group dance setting. $\downarrow$ indicates lower is better, $\uparrow$ indicates higher is better. \textbf{Best} and \underline{runner-up} scores are highlighted.}
\footnotesize
\setlength{\tabcolsep}{3.5pt}

\begin{tabular}{lcccccccccc}
\toprule

\multirow{2}{*}{Method}
& \multicolumn{3}{c}{Group Dance}
& \multicolumn{3}{c}{Single Dance}
& \multicolumn{4}{c}{Efficiency} \\

\cmidrule(lr){2-4}
\cmidrule(lr){5-7}
\cmidrule(lr){8-11}

& GMR$\downarrow$ 
& GMC$\uparrow$ 
& TIF$\downarrow$
& FID$\downarrow$ 
& Div$\uparrow$ 
& PFC$\downarrow$
& \makecell{FLOPs\\(G)}
& \makecell{Params\\(M)}
& \makecell{Train\\(min/epoch)}
& \makecell{Inf\\(s)} \\

\midrule

EDGE & 67.24 & 57.65 & 0.38 & 35.40 & 7.97 & 3.87 & 68.21  & 55.27 & 2:56 & 104.6 \\
GCD & 40.68 & 79.25 & 0.28 & 51.25 & 7.24 & 3.52 & 92.93  & 64.42 & 4:12 & 92.4  \\
CoDancers & 35.20 & 70.53 & \textbf{0.14} & 43.62 & 5.48 & 4.23 & 153.70 & 59.32 & 5:23 & 72.3  \\
TCDiff++ & \underline{14.67} & \underline{81.64} & \underline{0.15} & \underline{22.47} & \underline{16.23} & \textbf{1.53} & 168.72 & 72.64 & 5:43 & 124.4 \\
ST-GDance & 34.25 & 74.42 & \underline{0.15} & 31.42 & 14.35 & 3.21 & \textbf{15.92}  & \textbf{51.42} & \textbf{1:31} & \textbf{30.5}  \\
ST-GDance++ & \textbf{14.32} & \textbf{82.47} & \textbf{0.14} & \textbf{20.37} & \textbf{16.19} & \underline{2.32} & \underline{29.32} & \underline{52.83} & \underline{1:45} & \underline{32.5} \\

\bottomrule
\end{tabular}
\label{tab_long}
\end{table*}

\textbf{Evaluating long-duration scenarios.}
To assess robustness under long-horizon generation, we extend the sequence length to 720 frames and report quantitative results in Table~\ref{tab_long}. As the generation horizon increases, all baselines exhibit noticeable performance degradation, manifested by reduced group coherence, accumulated temporal drift, and increased collision frequency. These issues become particularly severe when spatial coordination or temporal consistency is not explicitly modeled. Single-dancer diffusion methods such as EDGE and GCD, which lack explicit spatial interaction modeling, tend to produce inconsistent dancer layouts over time, leading to frequent position swaps and degraded group formation realism. Although EDGE maintains relatively stable individual motion quality, the absence of global structural constraints results in more collisions and poorer group-level metrics. CoDancers focuses primarily on individual-level synthesis without sufficiently capturing inter-dancer dependencies, which weakens global coordination. ST-GDance improves temporal smoothness through spatial–temporal factorization, yet its autoregressive decoding accumulates errors over long sequences, causing gradual drift and reduced stability. TCDiff++ achieves stronger coherence by jointly modeling positions and motions in an end-to-end manner; however, its dense temporal modeling introduces higher computational complexity and still struggles to fully eliminate long-horizon inconsistencies. In contrast, our ST-GDance++ consistently achieves the best or second-best performance across nearly all metrics in long-duration settings. By explicitly decoupling spatial and temporal modeling and adopting segment-wise parallel diffusion, our approach preserves intra-group consistency while preventing long-term error accumulation. Consequently, it produces more coherent formations (lower GMR), stronger inter-dancer correlation (higher GMC), fewer collisions (lower TIF), and richer motion diversity (higher Div), demonstrating superior robustness and scalability for long-duration group choreography generation.

\begin{table}[tb]
\centering
\caption{Quantitative comparison of different methods on group dance generation with varying numbers of dancers.}
\label{tab:dance_comparison}
\begin{tabular}{lcccc}
\toprule
Method & Dancers & GMR$\downarrow$ & GMC$\uparrow$ & TIF$\downarrow$ \\
\midrule
\multirow{4}{*}{GCD} 
& 2 & 34.39 & 80.32 & 0.17 \\
& 3 & 30.22 & 80.22 & 0.19 \\
& 4 & 36.28 & 81.82 & 0.13 \\
& 5 & 38.43 & 81.44 & 0.17 \\
\midrule
\multirow{4}{*}{CoDancers} 
& 2 & 24.55 & 72.52 & \textbf{0.08} \\
& 3 & 26.34 & 74.22 & \textbf{0.10} \\
& 4 & 26.44 & 75.34 & \textbf{0.10} \\
& 5 & 27.27 & 74.34 & \textbf{0.11} \\
\midrule
\multirow{4}{*}{TCDiff++} 
& 2 & \underline{14.52} & \underline{81.46} & \textbf{0.08} \\
& 3 & \underline{11.85} & \underline{82.04} & \underline{0.11} \\
& 4 & \underline{12.42} & \textbf{82.92} & \underline{0.11} \\
& 5 & \underline{12.95} & \underline{81.34} & 0.13 \\
\midrule
\multirow{4}{*}{ST-GDance} 
& 2 & 19.42 & 80.52 & 0.12 \\
& 3 & 14.76 & 81.24 & 0.11 \\
& 4 & 14.02 & 81.42 & \textbf{0.10} \\
& 5 & 23.22 & 80.76 & \underline{0.12} \\
\midrule
\multirow{4}{*}{\textbf{ST-GDance++}} 
& 2 & \textbf{13.01} & \textbf{81.52} & \underline{0.11} \\
& 3 & \textbf{10.92} & \textbf{82.83} & \textbf{0.10} \\
& 4 & \textbf{12.06} & \underline{82.46} & 0.12 \\
& 5 & \textbf{12.75} & \textbf{81.86} & \underline{0.12} \\
\bottomrule
\end{tabular}
\label{tab_varnums}
\footnotetext{$\downarrow$ indicates lower is better, while $\uparrow$ indicates higher is better. \textbf{Best} and \underline{runner-up} scores are highlighted.}
\end{table}

\textbf{Efficiency Analysis.}
As shown in Tables~\ref{tab_short} and~\ref{tab_long}, methods built on standard Transformer-style sequence modeling exhibit a clear scaling bottleneck when extending the generation horizon. This is most evident for TCDiff++ and GCD: their FLOPs increase sharply from the short setting to the long setting (e.g., TCDiff++ from 32.56 to 168.72 GFLOPs; GCD from 27.69 to 92.93 GFLOPs), and the training/inference latency grows accordingly (TCDiff++ reaches 5:43 per epoch and 124.4 s for inference in the long setting). Such growth is consistent with the quadratic attention cost with respect to sequence length, making long-duration generation substantially more expensive even when batch size and dancer count remain unchanged.

CoDancers, in contrast, follows a VAE-style formulation and benefits from relatively faster inference in the short setting, but its overall cost still scales aggressively in the long setting (53.95 to 153.70 GFLOPs), and its efficiency advantage is less stable once the sequence becomes long. This suggests that while VAE-based designs can reduce per-step sampling overhead, they may still suffer from heavy temporal decoding and reduced scalability when the target horizon is extended.

ST-GDance and ST-GDance++ are designed specifically to improve long-horizon efficiency within the diffusion framework. Comparing ST-GDance and ST-GDance++ in Tables~\ref{tab_short} and~\ref{tab_long} shows a clear trade-off. ST-GDance is highly efficient, requiring only 6.75 GFLOPs and 0:37 training time in the short setting, and remaining the most lightweight method in the long setting. However, its generation quality is slightly behind the best-performing diffusion models. But ST-GDance++ improves these metrics while keeping the model compact. It achieves the best group coordination scores in both short and long settings, and maintains strong diversity. At the same time, its computational cost remains moderate, with 29.32 GFLOPs and 1:45 training time in the long setting, far lower than TCDiff++. This indicates that ST-GDance++ achieves stronger generation quality while still maintaining high efficiency.

\begin{table*}[t]
\centering
\caption{Ablation study of ST-GDance++ on different components. SMB, AAM, and TNS denote the Spatial Modeling Block, Alignment Attention Mask, and Triangular Noise Schedule, respectively. \textbf{Best} results are highlighted in bold.}
\label{tab:ablation_study}
\footnotesize 
\begin{tabular*}{\textwidth}{@{\extracolsep\fill}ccc|ccc|ccc|cccc}
\toprule
\multicolumn{3}{c|}{Method} & \multicolumn{3}{c|}{Group Dance} & \multicolumn{3}{c|}{Single Dance} & \multicolumn{4}{c}{Efficiency} \\
\cmidrule(lr){1-3} \cmidrule(lr){4-6} \cmidrule(lr){7-9} \cmidrule(lr){10-13}
SMB & AAM & TNS & GMR$\downarrow$ & GMC$\uparrow$ & TIF$\downarrow$ & FID$\downarrow$ & Div$\uparrow$ & PFC$\downarrow$ & \makecell{FLOPs\\(G)} & \makecell{Params\\(M)} & \makecell{Train\\(min/ep)} & \makecell{Inf\\(s)} \\
\midrule
\checkmark & \checkmark & & 12.53 & 81.49 & 0.11 & 24.53 & 18.02 & 1.44 & \textbf{11.32} & \textbf{51.64} & \textbf{0:42} & \textbf{4.2} \\
\checkmark & & \checkmark & 14.32 & 80.57 & 0.12 & 25.52 & 17.94 & 1.52 & 14.63 & 58.42 & 0:53 & 5.2 \\
 & \checkmark & \checkmark & 15.45 & 79.43 & 0.13 & 25.45 & 18.54 & 1.38 & 30.42 & 68.38 & 1:27 & 8.4 \\
\midrule
\multicolumn{3}{l|}{ST-GDance++(Full)} & \textbf{10.85} & \textbf{82.83} & \textbf{0.10} & \textbf{25.34} & \textbf{18.76} & \textbf{1.24} & \textbf{11.32} & \textbf{51.64} & \textbf{0:42} & \textbf{4.2} \\
\bottomrule
\end{tabular*}
\end{table*}

\textbf{Comparison of Multi-Dancer Generation.} To evaluate the robustness of our model in handling different crowd densities, we report the quantitative results for varying numbers of dancers (from 2 to 5) in Table~\ref{tab_varnums}. 
Our proposed \textbf{ST-GDance++} consistently achieves the state-of-the-art performance across nearly all metrics and configurations. 
Notably, as the complexity of group dynamics increases with the number of dancers, baseline methods such as CoDancers exhibit a significant performance degradation; for instance, its GMR increases sharply from 24.55 to 27.27 when scaling from 2 to 5 dancers. In contrast, our model maintains a remarkably stable and superior GMR (12.75 for $N=5$), demonstrating its effective modeling of complex spatio-temporal dependencies among multiple agents. Furthermore, ST-GDance++ leads in Group Motion Consistency (GMC) across all settings (e.g., reaching 82.83 for $N=3$), which underscores its capability to generate highly synchronized and harmonious group choreographies. While TCDiff++ and CoDancers show competitive results in Trajectory Intersection Frequency (TIF) in some cases, they often do so at the cost of motion fidelity or coordination. Our approach strikes a superior balance, achieving the best GMR and GMC while maintaining a comparable collision avoidance rate. 

\subsection{Ablation Study} \label{ablation}

\textbf{Impact of Triangular Noise Schedule (TNS)}
The Triangular Noise Schedule (TNS) primarily affects the quality of generated sequences. As shown in Table \ref{tab:ablation_study}, when TNS is applied, both group dance and single dance generation performance improve. Specifically, the GMR score decrease from 12.53 to 10.85, which represents a 13.4\% reduction. Additionally, diversity (Div) increases from 18.02 to 18.76, marking a 4.1\% improvement. During training, we apply a varying noise schedule for different segments of the sequence, which introduces more complexity and variability, helping the model learn to handle a wider range of noisy scenarios. In contrast, during testing, we use the traditional uniform noise addition method, which simplifies the noise schedule. This discrepancy makes the training process more challenging, but it also increases the model's robustness, as it is exposed to more diverse noise patterns. Since the introduction of TNS does not modify the network architecture, it does not introduce any additional computational overhead during either training or inference. Therefore, the efficiency-related metrics remain unchanged. The increased robustness during training translates into more stable and higher-quality outputs during testing, without affecting the efficiency of inference.

\begin{figure*}[tb]
    \centering
    \includegraphics[width=0.9\linewidth]{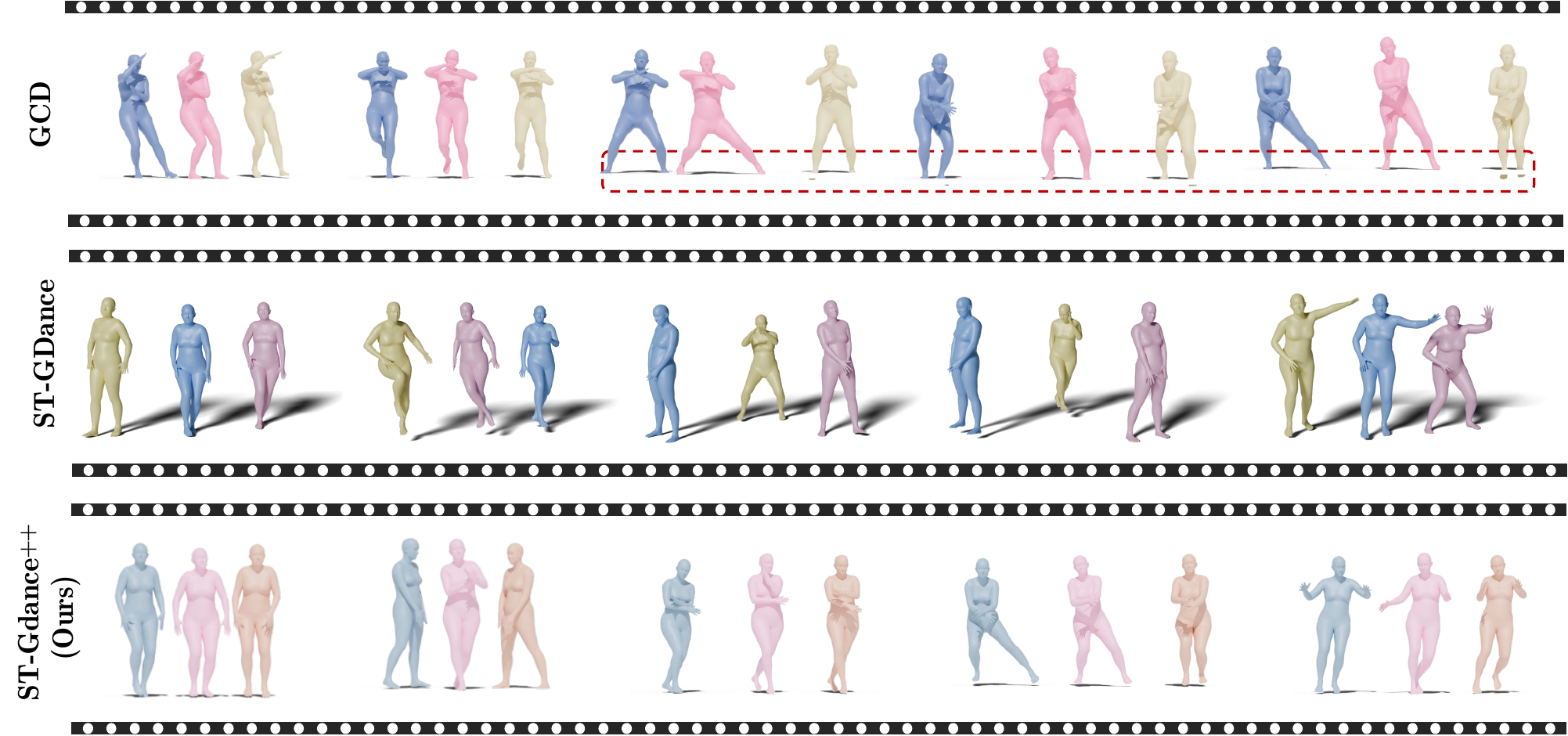}
    \caption{Qualitative results compared with baselines.}
\label{fig:baseline}
\end{figure*}

\textbf{Impact of Spatial Modeling Block (SMB)}
The inclusion of the Spatial Modeling Block (SMB) significantly improves the performance of group dance generation. As shown in the table, when SMB is added, the GMR decreases from 15.45 to 10.85, which is a 29.8\% reduction, indicating that the group motion generation becomes more coherent. Similarly, the TIF score reduces from 0.13 to 0.10, showing a 23.1\% reduction in trajectory intersection frequency. These improvements highlight the ability of SMB to enhance spatial coordination among dancers in group settings. However, for single dance generation, the impact of SMB is minimal. Thus, SMB primarily benefits group dance generation.

\textbf{Impact of Alignment Attention Mask (AAM)}
The use of the Alignment Attention Mask (AAM) improves both group and single dance generation quality, with a more noticeable effect in group scenarios. As shown in Table \ref{tab:ablation_study}, configurations with AAM consistently achieve better motion consistency and alignment, leading to improved group coordination metrics. In the single dance setting, AAM also contributes to lower FID and more stable motion patterns. In terms of efficiency, AAM also helps to decrease model complexity by removing future attention calculations. The training time per epoch decrease from 0:53 to 0:42, corresponding to a 20.8\% decrease, and the number of parameters decrease from 58.42 M to 51.64 M. Overall, these results show that AAM improves efficiency while retaining motion alignment and generation quality.

\subsection{Qualitative Visual Comparison}
\label{qualitative}

\textbf{Baseline comparison.}
We qualitatively compare ST-GDance++ with representative baselines, including GCD and the original ST-GDance. As shown in Fig.~\ref{fig:baseline}, each row presents the generated motion sequences of different methods under the same music condition.

GCD tends to produce noticeable foot penetration artifacts, where the feet unrealistically intersect with the ground or exhibit abrupt positional shifts. This issue becomes more apparent in sequences involving fast or complex steps, indicating that the method struggles to maintain stable contact constraints and consistent spatial coordination among dancers. The original ST-GDance alleviates some spatial coordination issues, but still suffers from foot sliding artifacts. In several frames, the feet appear to drift along the ground instead of maintaining stable contact, resulting in visually unnatural motion. This phenomenon suggests that, although ST-GDance captures overall group structure, its temporal modeling is insufficient for fine-grained motion accuracy, especially in long or complex sequences.

In contrast, ST-GDance++ generates more stable and physically plausible motions. The feet maintain consistent ground contact, and the transitions between steps are smoother and more natural. Moreover, inter-dancer spacing remains coherent throughout the sequence, with fewer collisions or abrupt adjustments. These visual improvements demonstrate the effectiveness of the proposed spatial modeling, alignment attention, and streaming-oriented temporal design in producing more realistic and coordinated group dance motions.

\begin{figure*}[tb]
    \centering
    \includegraphics[width=0.9\linewidth]{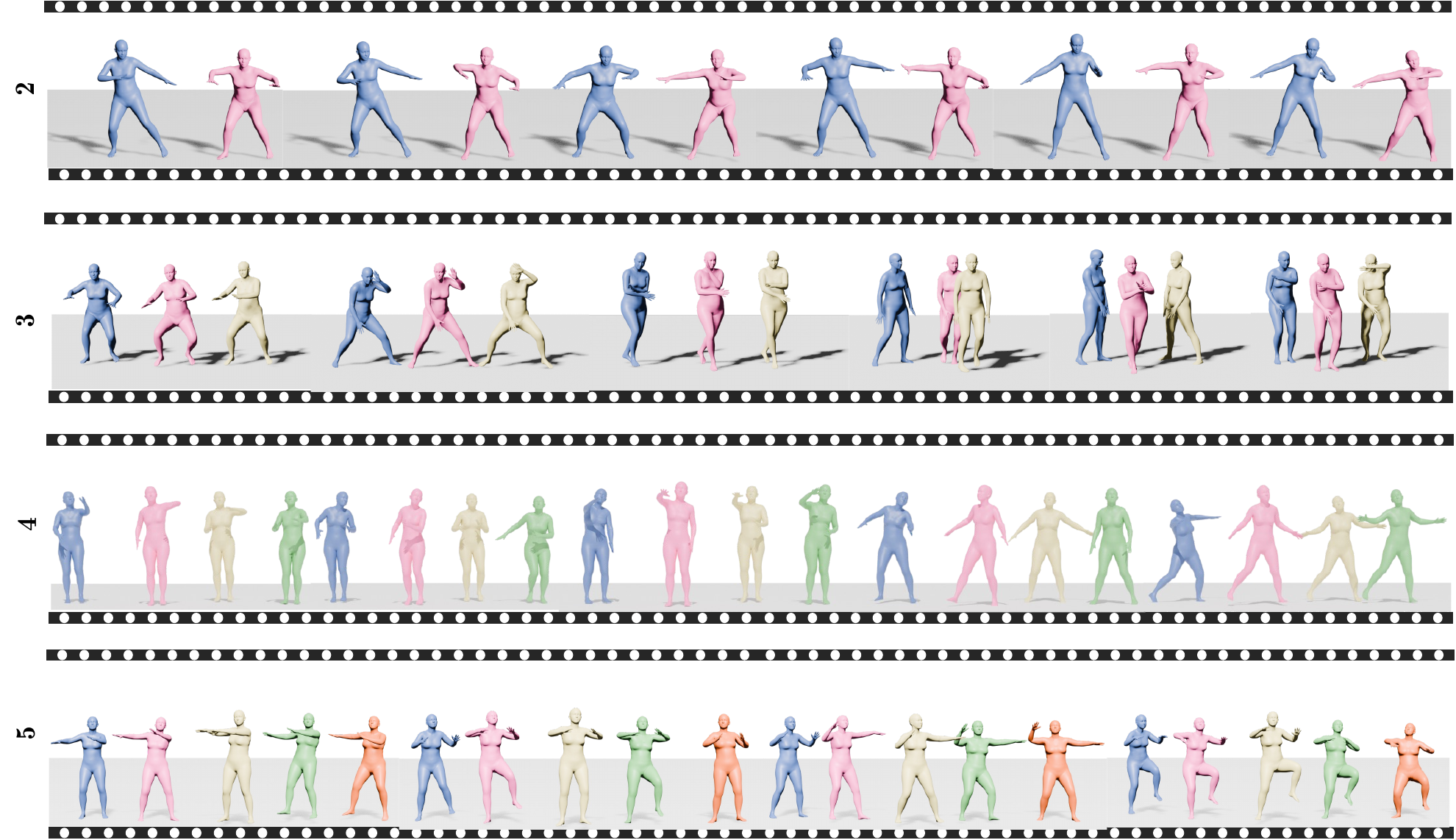}
    \caption{Qualitative results generated by ST-GDance++ across varying group sizes. We demonstrate the generated group choreographies with different numbers of dancers, ranging from 2 to 5.}
\label{fig:nums}
\end{figure*}

\textbf{Varying Group Sizes.} We further evaluate the scalability of ST-GDance++ by testing the model under different group sizes, ranging from two to five dancers. As illustrated in Fig.~\ref{fig:nums}, ST-GDance++ consistently produces stable and coordinated motions across all group configurations. When the number of dancers increases, the spatial layout becomes more crowded and the interaction complexity grows. Despite this, the generated motions remain well-organized, with consistent inter-dancer spacing and synchronized movement patterns. Even in the five-dancer setting, the model avoids collisions and maintains coherent formations throughout the sequence. This result demonstrates that ST-GDance++ scales effectively with the number of dancers. The explicit spatial modeling mechanism allows the model to capture inter-dancer relationships without relying on dense attention over concatenated tokens, preventing the quadratic growth of computational complexity. As a result, the model preserves both motion quality and spatial consistency as the group size increases, highlighting its suitability for long-duration and large-group choreography generation.

\begin{figure}[tb]
    \centering
    \includegraphics[width=0.9\linewidth]{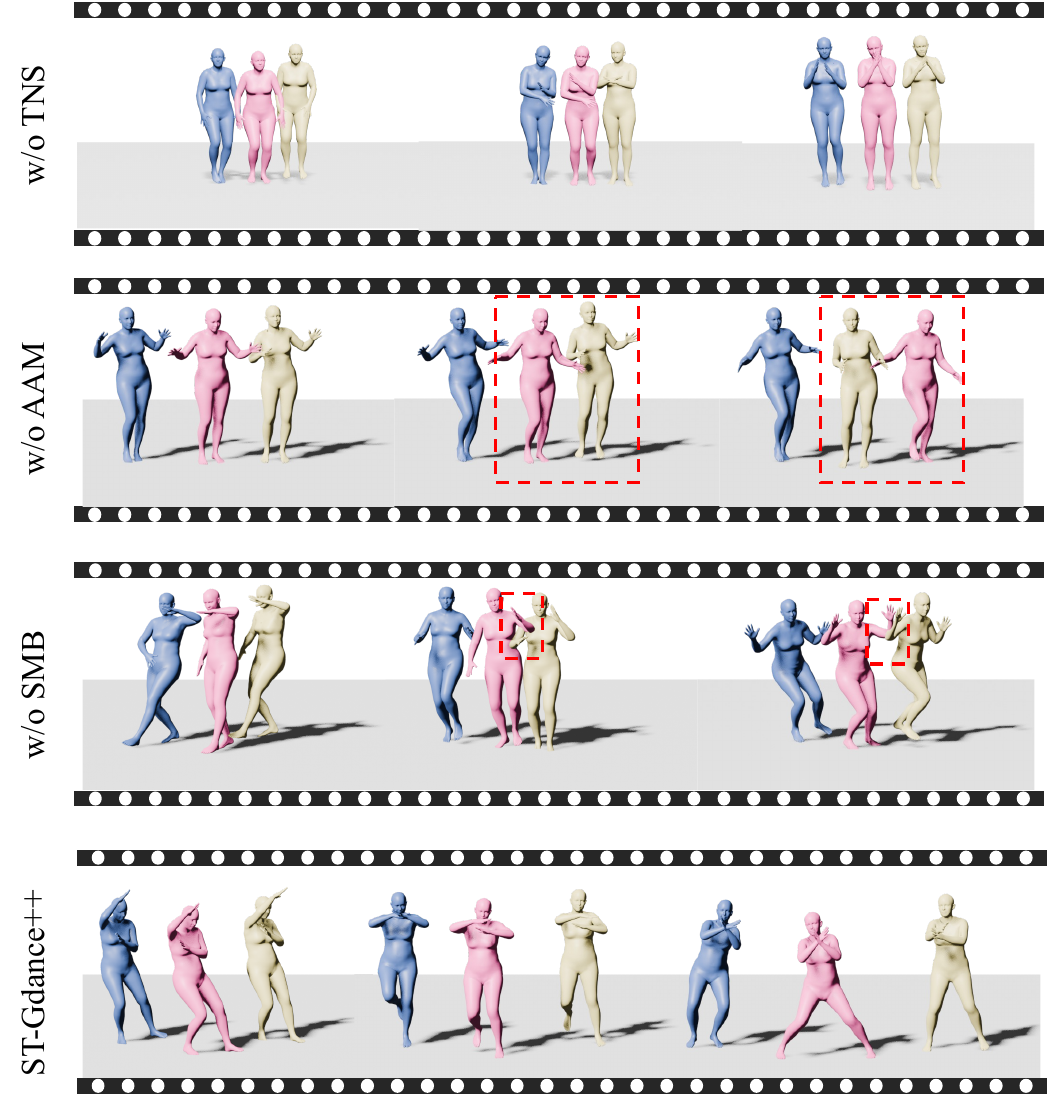}
    \caption{Qualitative comparison under different component ablations of ST-GDance++.}
\label{fig:abla}
\end{figure}

\textbf{Visualization of the ablation experiment.}
We further present qualitative comparisons of different ablated variants of ST-GDance++ to analyze the effect of each component. As shown in Fig.~\ref{fig:abla}, removing each module leads to distinct motion artifacts. 

Without the Triangular Noise Schedule (w/o TNS), the generated motions become overly similar across dancers, with reduced diversity in poses and transitions. This phenomenon is consistent with the quantitative results in Table~\ref{tab:ablation_study}, where the diversity metric drops compared to the full model, indicating that TNS contributes to richer motion variations. When the Alignment Attention Mask is removed (w/o AAM), we observe abrupt swapping behaviors between dancers, as highlighted in the red dashed regions. The dancers occasionally exchange spatial positions in an unnatural manner, leading to inconsistent temporal alignment. This visual artifact corresponds to the degradation in group motion metrics reported in Table~\ref{tab:ablation_study}, suggesting that AAM is essential for maintaining stable inter-dancer alignment over time. Without the Spatial Modeling Block (w/o SMB), the generated sequences exhibit frequent collisions, particularly around the arms and upper body. As shown in the highlighted regions, dancers’ limbs intersect due to the lack of explicit spatial interaction modeling. This observation is consistent with the increased GMR and TIF values in Table~\ref{tab:ablation_study}, indicating poorer spatial coordination.

In contrast, the full ST-GDance++ model produces stable, well-coordinated motions with clear inter-dancer spacing, smooth transitions, and diverse poses. These qualitative results, together with the quantitative improvements in Table~\ref{tab:ablation_study}, demonstrate that each component plays a complementary role in improving motion diversity, alignment, and spatial consistency.

\section{Limitation}
\label{limitate}

Despite the improvements achieved by ST-GDance++ in long-duration group dance generation, several limitations remain. First, the current framework is designed for a single cross-modal setting, where dance motions are generated solely from music input. It does not yet support more flexible or user-driven controls, such as textual descriptions, formation constraints, or motion-level editing signals. While focusing on this basic setting allows us to study the core spatial–temporal generation problem, more practical applications would require richer and more interactive control mechanisms. Second, although the proposed spatial modeling improves overall coordination, the model still struggles with certain complex interaction patterns, such as rapid role exchanges or formation swaps among dancers. These behaviors are inherently difficult to model, and are also underrepresented in existing datasets, which typically contain limited examples of such interactions. As a result, the model’s performance in highly dynamic group choreography remains constrained. Future work will explore incorporating richer control signals and improving the modeling of complex interaction patterns, as well as leveraging more diverse and better-annotated datasets for large-scale group choreography generation.

\section{Conclusion}

In this work, we presented ST-GDance++, a scalable spatial–temporal diffusion framework for long-duration group dance generation. By decoupling spatial interaction modeling from temporal motion dynamics, the proposed method avoids the quadratic complexity of dense attention over dancer–frame tokens. Specifically, we introduce a lightweight spatial modeling block to explicitly capture inter-dancer relationships, an alignment-aware attention mechanism for stable coordination, and a streaming-oriented noise schedule to support long-sequence generation. Extensive experiments demonstrate that ST-GDance++ achieves improved motion quality, spatial consistency, and computational efficiency across both single and group dance scenarios. Qualitative and quantitative results show that the proposed components complement each other, leading to more stable, diverse, and well-coordinated motions, especially in long-duration and multi-dancer settings.

\backmatter

\bmhead{Supplementary information}

The supplementary material includes (i) the original conference version of this work (ST-GDance) for reference and (ii) additional qualitative results in the form of generated multi-dancer dance videos.

\section*{Declarations}
\bmhead*{Conflict of interest}
The authors declare that they have no known competing financial
interests or personal relationships that could have appeared to influence the work reported in this paper.

\bmhead*{Data availability}
The experiments in this study are conducted using the publicly available AIOZ-GDance dataset~\citep{aiozGdance}, which can be accessed at \url{https://huggingface.co/datasets/aiozai/AIOZ-GDANCE}.




\begin{appendices}





\end{appendices}

\bibliography{sn-bibliography}

\end{document}